\documentclass[letterpaper, 10 pt, conference]{ieeeconf}  

\IEEEoverridecommandlockouts                              
                                                 
\usepackage{graphics} 
\usepackage{epsfig} 
\usepackage{mathptmx} 
\usepackage{times} 
\usepackage{amsmath} 
\usepackage{amssymb}  
\usepackage{url}
\usepackage{romannum}
\usepackage{xcolor}
\usepackage{float}
\usepackage{url}
\usepackage[ruled,vlined]{algorithm2e}
\usepackage{array}
\usepackage{placeins}

\title{\LARGE \bf
Sequence-Based Filtering for Visual Route-Based Navigation: Analysing the Benefits, Trade-offs and Design Choices
}

\author{Mihnea-Alexandru Tomiță$^{1}$, Mubariz Zaffar$^{2}$, Michael Milford$^{3}$, Klaus McDonald-Maier$^{1}$ and Shoaib Ehsan$^{1}$
\thanks{$^{1}$Authors are with the School of Computer Science and Electronic Engineering,
        University of Essex, CO4 3SQ, United Kingdom
        {\tt\small matomi@essex.ac.uk, kdm@essex.ac.uk, sehsan@essex.ac.uk}. }%

\thanks{$^{2}$Mubariz Zaffar is with the Cognitive Robotics Department, Delft University of Technology, Delft, CD 2628, Netherlands
        {\tt\small m.zaffar@tudelft.nl}}%

\thanks{$^{3}$Michael Milford is with the School of Electrical Engineering and Computer Science, Queensland University of Technology, Brisbane, QLD 4000, Australia
        {\tt\small michael.milford@qut.edu.au}}%
        
\thanks{This work is supported by the UK Engineering and Physical Sciences Research Council through grants EP/R02572X/1 and EP/P017487/1.}
}

\begin{document}

    \maketitle
    \thispagestyle{empty}
    \pagestyle{empty}

    \begin{abstract}
    Visual Place Recognition (VPR) is the ability to correctly recall a previously visited place using visual information under environmental, viewpoint and appearance changes. An emerging trend in VPR is the use of sequence-based filtering methods on top of single-frame-based place matching techniques for route-based navigation. The combination leads to varying levels of potential place matching performance boosts at increased computational costs. This raises a number of interesting research questions: How does performance boost (due to sequential filtering) vary along the entire spectrum of single-frame-based matching methods? How does sequence matching length affect the performance curve? Which specific combinations provide a good trade-off between performance and computation? However, there is lack of previous work looking at these important questions and most of the sequence-based filtering work to date has been used without a systematic approach. To bridge this research gap, this paper conducts an in-depth investigation of the relationship between the performance of single-frame-based place matching techniques and the use of sequence-based filtering on top of those methods. It analyzes individual trade-offs, properties and limitations for different combinations of single-frame-based and sequential techniques. A number of state-of-the-art VPR methods and widely used public datasets are utilized to present the findings that contain a number of meaningful insights for the VPR community.
    \end{abstract}


    \section{Introduction}\label{introduction}
      The goal of a visual place recognition (VPR) system is to determine if a currently observed place has been previously visited by a robot/human. In recent years, it has been shown that sequence-based VPR systems such as \cite{milford2012seqslam, milford2013vision, chancan2020deepseqslam, vysotska2015lazy} and \cite{tomitua2020convsequential} can achieve good performance in changing environments. Thus, an almost parallel track has emerged where sequence-based techniques has been shown to outperform single-frame-based techniques. More importantly, the benefits presented by sequential information are generally extendable to most non-learning and learning-based VPR techniques albeit at varying levels and costs. Therefore, it is critical to understand the properties of sequential-based filtering, its trade-offs and how to deploy them on single-frame-based VPR techniques for designing better VPR systems.   
      \endgraf

      To the best of our knowledge, there is no previous work that has examined this important problem in a systematic way (such as performance boost variations due to sequential filtering along the entire spectrum of single-frame-based VPR methods, the effects of sequence length on performance, performance-computation trade-off etc). To bridge this research gap, this paper investigates the relationship between the performance of single-frame-based, learnt and non-learnt VPR methods, and the use of sequence-based filtering on top of these methods. In particular, this paper introduces sequential information into a number of VPR techniques to improve conditional invariance and show that sequence matching takes a poorly performing single-frame-based VPR technique and improves its performance. The paper also highlights that for a high precision system, adding sequential matching tends to dilute the differential effects of single-frame matching systems. It examines the effects of different sequence lengths on the resulting performance boost and determines the most optimal combinations between different VPR techniques and sequence lengths, taking into consideration both the performance and computational load of each system.
  
      The remainder of this paper is organised as follows: Section \ref{literature_review} presents an overview of the literature regarding VPR. Section \ref{methodology} presents our implementation of sequential-based filtering on top of single-frame-based methods. Section \ref{experimental_study} describes the experimental setup for performing the analysis on trade-offs of sequential filtering for VPR. Section \ref{results} presents the detailed results and analysis. Finally, the conclusions are presented in Section \ref{conclusion}.

    \section{Literature Review}
    \label{literature_review}
    
    
        Early techniques used in the field of VPR were based on handcrafted feature descriptors \cite{bay2006surf}, \cite{lowe2004distinctive}. Scale-Invariant Feature Transform (SIFT) \cite{SIFT} and Speeded-Up Robust Features (SURF) \cite{SURF} have been used to solve VPR problem such as in \cite{murillo2007surf}, \cite{stumm2013probabilistic}, \cite{andreasson2004topological}, \cite{kovsecka2005global} and \cite{se2002mobile}. Gist \cite{oliva2006building}, \cite{oliva2001modeling} has been used in \cite{murillo2009experiments}, \cite{singh2010visual} and \cite{sunderhauf2011brief} for image matching. BRIEF has been paired with Gist by the authors of \cite{sunderhauf2011brief}. Histogram-of-Oriented-Gradients (HOG) \cite{freeman1995orientation}, \cite{dalal2005histograms} is another whole-image descriptor used by the authors of \cite{mcmanus2014scene}. Zaffar \textit{et al.} \cite{zaffar2020cohog} employ HOG feature descriptors to achieve state-of-the-art place matching performance. \endgraf
        
        Convolutional Neural Networks (CNNs) have been widely explored by researchers (such as in \cite{sharif2014cnn} and \cite{oquab2014learning}) in VPR. Chen \textit{et al.} \cite{chen2014convolutional} used the spatial filter of SeqSLAM together with all the layers of the Overfeat Network \cite{sermanet2013overfeat}. The authors of \cite{chen2017deep} created two neural-network based VPR techniques. The first architecture, entitled HybridNet, used weights learnt from the top 5 convolutional layers of CaffeNet \cite{krizhevsky2012imagenet}, while the second architecture, AMOSNet, was trained from scratch on the SPED dataset. The authors of NetVLAD \cite{arandjelovic2016netvlad} presented a new Vector-of-Locally-Aggregated-Descriptors (VLAD) layer that can be incorporated in any neural network architecture, drastically enhancing the performance in VPR related scenarios. Merrill \textit{et al.} \cite{merrill2018lightweight} showed that convolutional auto-encoders are suitable for VPR tasks. The resulting CNN, namely CALC, is lightweight as well as robust to variations in both illumination and viewpoint. \endgraf
        
        SeqSLAM \cite{milford2012seqslam} performs visual place recognition in changing environments by comparing sequences of camera frames. SMART \cite{pepperell2014all} extended SeqSLAM by incorporating the varying speed of a vehicle. The authors of \cite{yang2020sequence} proposed a new sequence-based VPR system for aerial robots. In \cite{garg2020fast}, sequence-based matching is used to resolve the collisions in the hash space. Johns \textit{et al.} \cite{johns2013feature} show a new method for appearance-based localisation, namely Feature Co-occurrence Maps. The authors of \cite{naseer2014robust} propose a system with robust localisation. More recently, in \cite{tomitua2020convsequential}, the authors have presented a sequence-based VPR system based on HOG descriptors. DeepSeqSLAM \cite{chancan2020deepseqslam} is a trainable CNN+RNN system that is successfully able to complete VPR related tasks in challenging environments.\endgraf
        
        
    
    \begin{figure}
            \centering
            \begin{tabular}{ c }
                \includegraphics[width=240pt]{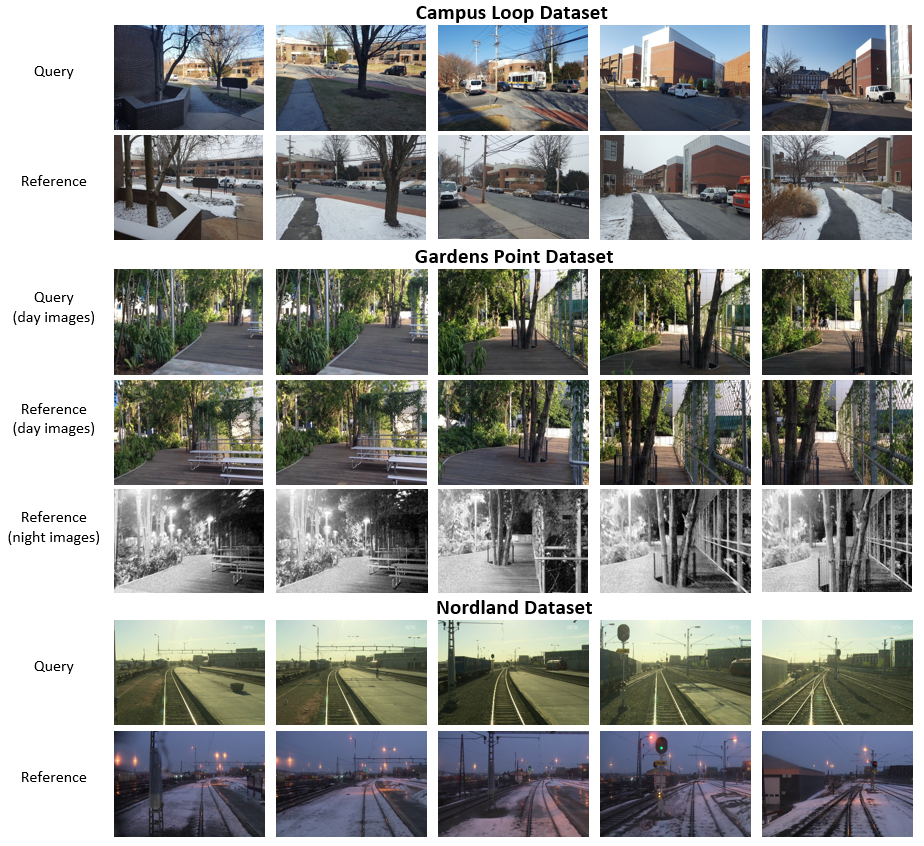}
            \end{tabular}
            \caption{Sample sequence of images taken from each of the 4 datasets: Campus Loop, Gardens Point (day-to-day), Gardens Point (day-to-night) and Nordland (summer-to-winter).  }
            \label{datasetimages}
        \end{figure}

    \section{Methodology} \label{methodology}
    This section presents the approach taken in this work for introducing sequence-based filtering on top of different single-frame-based VPR techniques.
     \subsection{Creating the Image Sequence}\label{imagesequence}
        The format of the 1D list containing query images is shown below: 
         
          \[ Sequential\ List = 
            \begin{bmatrix}
                q_{1} & q_{2} & q_{3} & \dots  & q_{K}
            \end{bmatrix}
            \]
         
        In the above equation, $q_1$ is the first query image, $q_K$ is the last query image, and \textbf{K} is the total number of images that forms each sequence. The following equation shows each 1D list, and the images that will be contained:  
        \medskip
        
        \[
            \begin{matrix}
                q_1 & q_2 & q_3 & \dots  & q_K \\
                q_2 & q_3 & q_4 & \dots  & q_{K+1} \\
                \vdots & \vdots & \vdots & \ddots & \vdots \\
                q_{N-K+1} & q_{N-K+2} & q_{N-K+3} & \dots  & q_{N}
            \end{matrix}
        \]
        \medskip
        
        As this technique matches consecutive images, only the first \textit{N - \textbf{K} + 1} images will be matched.\endgraf
        \medskip
        
         \begin{algorithm}
            \SetAlgoLined
                \textit{Given:} Total Number of Query Images \\
                \textit{Given:} Total Number of Reference Image \\
                \textbf{K} = image sequence length \\
                \For{i \textbf{in range} (total\_Query\_Images - \textbf{K} + 1)}{
                ref\_matching\_scores = [] \\
                \For{j \textbf{in range} (total\_Ref\_Images - \textbf{K} + 1)}{
                score = \textit{perform\_VPR($Q_F$, $R_F$, i, j)} \\
                ADD score to ref\_matching\_scores \\
                }
                }
                Best Match = Max (ref\_matching\_scores)
                
            \caption{Matching Query and Reference Sequences}
        \end{algorithm}
        
        \subsection{Sequence Matching}
         Query and reference features are initially computed and stored separated in two 1D lists ($Q_{F}$ and $R_{F}$). The function \textit{perform\_VPR} in Algorithm 1 is responsible for matching the sequential 1D query and reference lists (see sub-section \ref{imagesequence}) in a 1-to-1 fashion. The \textit{perform\_VPR} function is also responsible for generating the matching scores between these sequences of images. The matching score of any query and reference sequences of images is calculated as the arithmetic mean of the matching scores of the pairs within these sequences. Thus, the sequence with the highest score is chosen as the best match.
         
    
    \section{Experimental Setup}
    This section discusses the employed performance metrics and the utilised VPR techniques and sequential datasets for this work.
    \label{experimental_study}
    
        
        \begin{table} 
            \caption{Feature encoding times of different VPR techniques.} 
            \begin{center}
               \begin{tabular}{ |c|c| } 
            
            \hline
            \textbf{VPR Technique} & \textbf{Feature Encoding Time (sec)}\\
            \hline
            AMOSNet & 0.36 \\
            \hline
            CALC & 0.027 \\
            \hline
            HOG & 0.0043 \\
            \hline
            HybridNet & 0.36\\
            \hline
            NetVLAD & 0.77 \\
            \hline
            \end{tabular}
            \end{center}
            \label{table:encodingtimes}
            \end{table}

        \subsection{Employed Performance Metrics}
        Area-under-the-Precision-Recall-Curve (AUC) is widely used in VPR research for evaluation purposes \cite{zaffar2020vpr}. Thus, it is also employed in this work utilizing (\ref{eq:precision}) and (\ref{eq:recall}):  
        
        \medskip
        \begin{equation}\label{eq:precision}
            Precision = \frac{True\ Positives}{True\ Positives + False\ Positives}
        \end{equation}    
        
        \begin{equation}\label{eq:recall}  
            Recall = \frac{True\ Positives}{True\ Positives + False\ Negatives}
        \end{equation}
        \smallskip
        
       The authors of \cite{merrill2018lightweight}, \cite{khaliq2019holistic}, \cite{zaffar2019state} and \cite{zaffar2019levelling} determined that the feature encoding time ($t_e$) of a VPR system to be an important performance indicator. In \cite{zaffar2020cohog}, the authors evaluated a system's performance using Performance-per-Compute-Unit (PCU). This is defined by combining precision at 100$\%$ recall (P$_{R100}$) with $t_e$ as in equation (\ref{eq:PCU}):
        
        \begin{equation}\label{eq:PCU}         
            PCU =  P_{R100}\ \times \ \log\bigg(\frac{t_{e\_max}}{t_e} + 9\bigg)
        \end{equation}
        
        In this equation, the maximum feature encoding time ($t_{e\_max}$) is used to represent the most resource intensive VPR technique, while $t_e$ represents the feature encoding times for each of the remaining techniques (where $t_e$ $<$ $t_{e\_max}$). It should be noted that without the scalar 9 in equation (\ref{eq:PCU}), the VPR technique with $t_e$ = $t_{e\_max}$ will always result in a PCU of 0. Techniques with higher precision and lower feature encoding time generally lie towards the higher spectrum of PCU, while compute-intensive and less precise techniques converge towards lower PCU values. Thereby, this addition provides a more interpretable range. Because PCU is a relative performance metric, it serves us value in this study.
        
        \begin{figure*}
            \centering
            \begin{tabular}{ c c }

                \includegraphics[width=155pt]{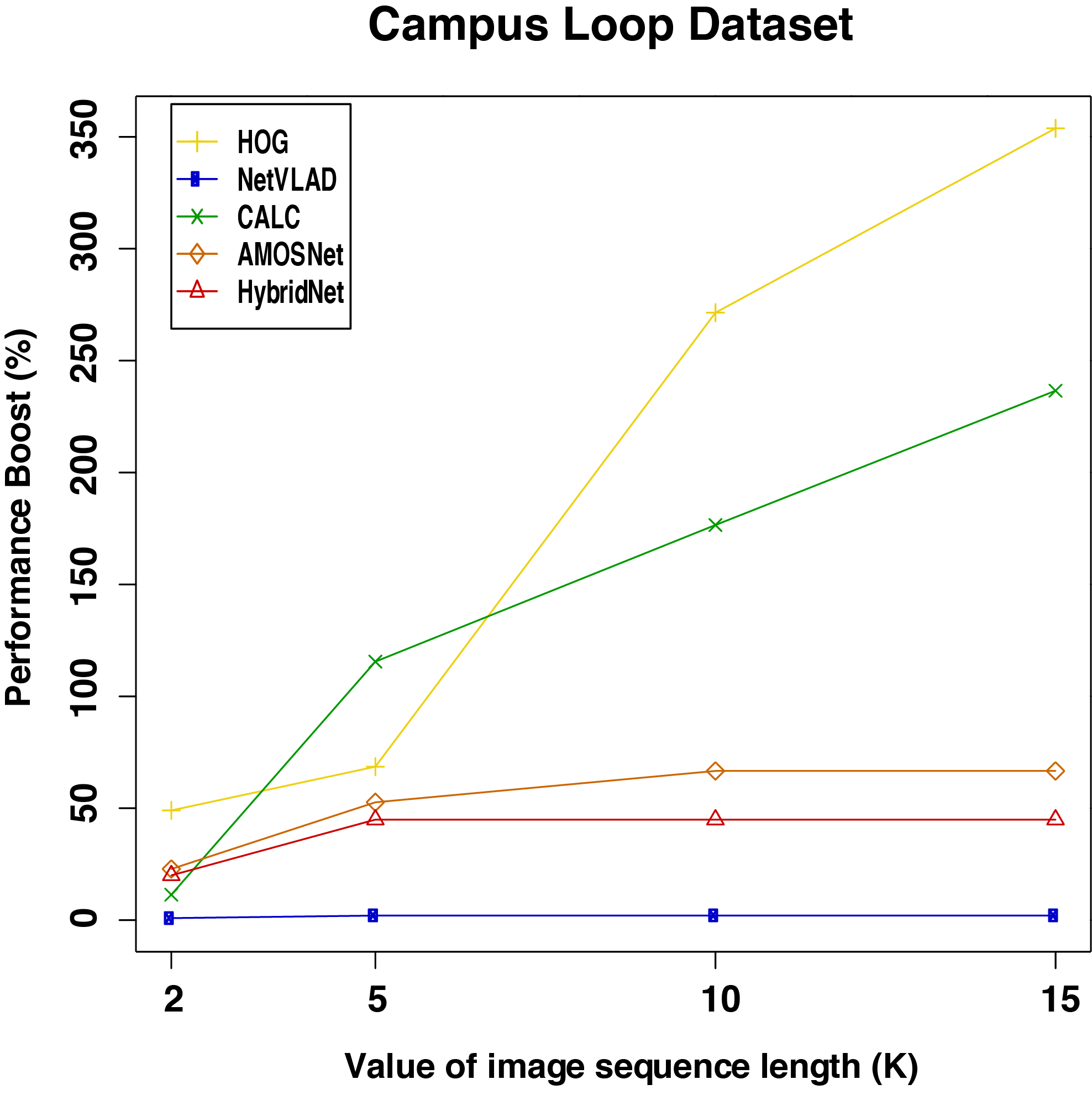} & 
                \includegraphics[width=155pt]{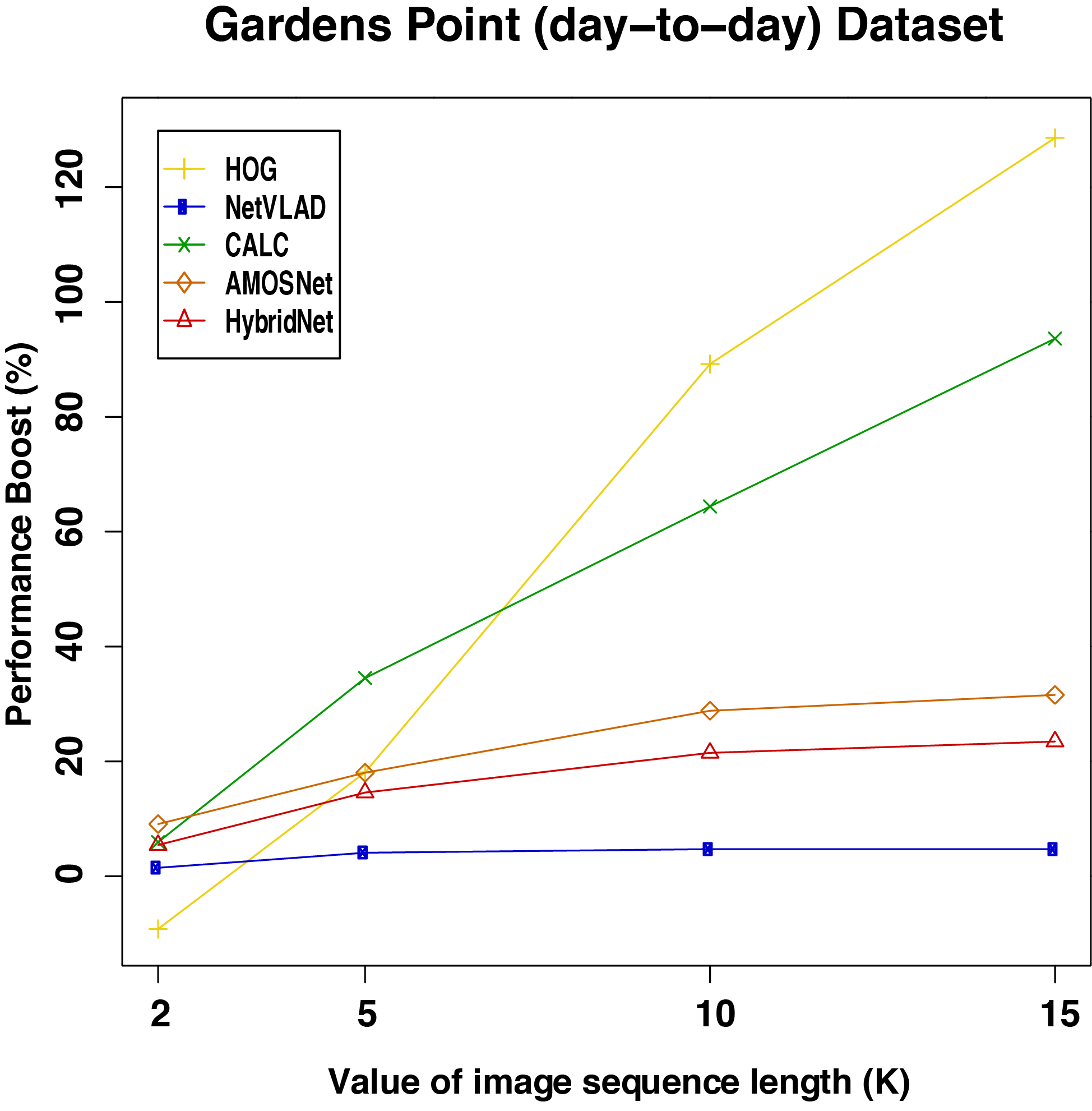} \\
                \includegraphics[width=155pt]{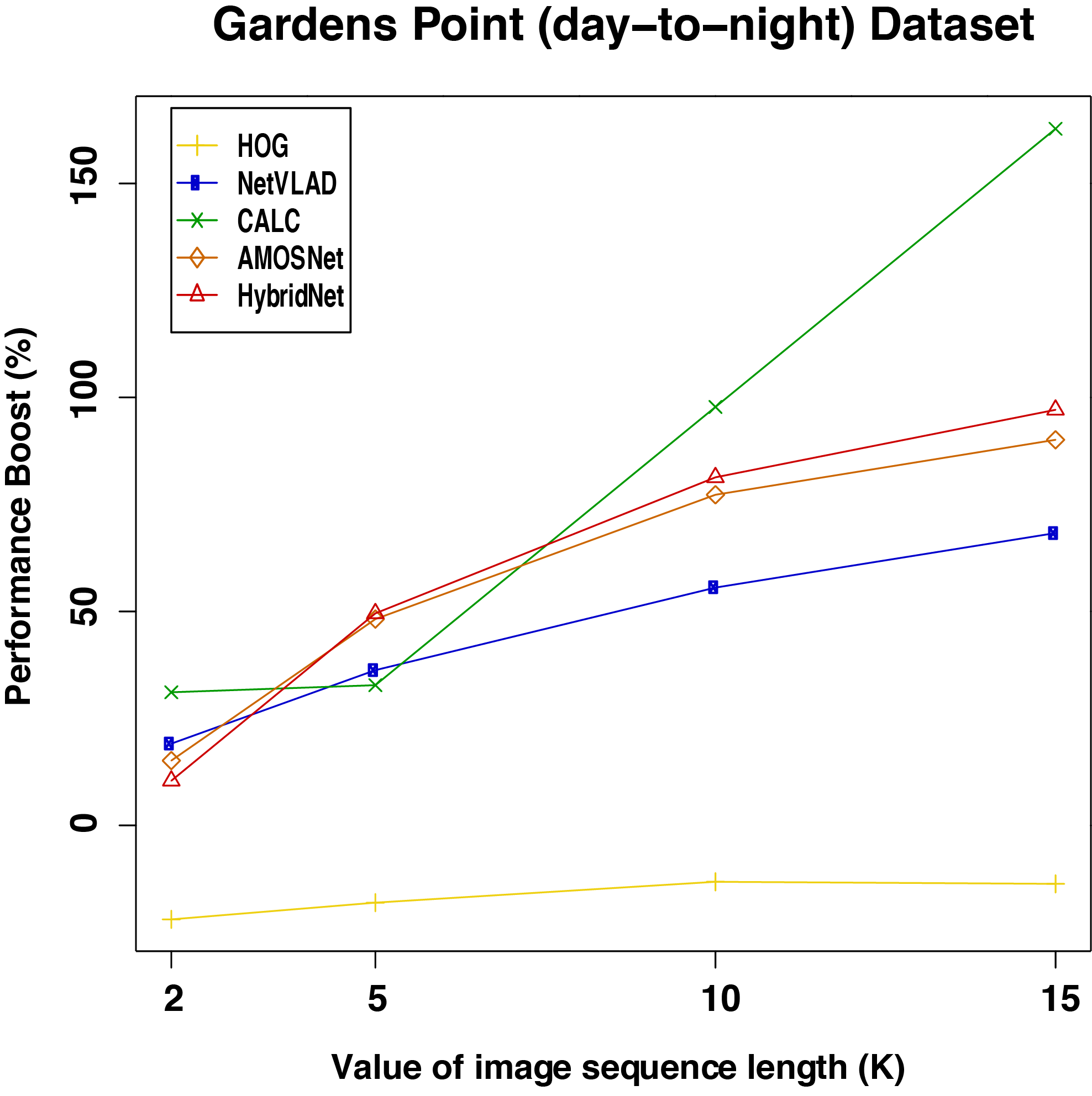}  &
                \includegraphics[width=155pt]{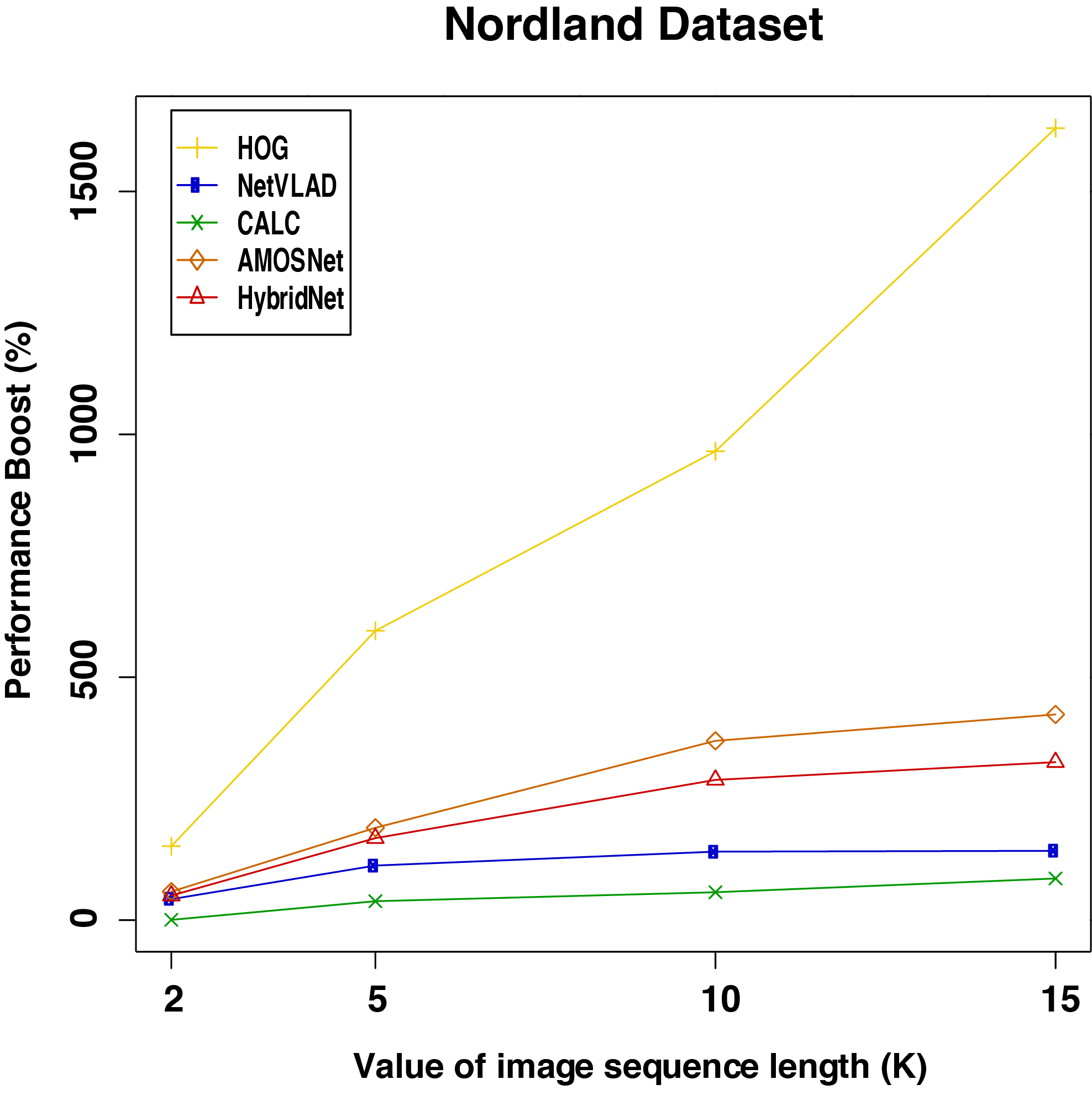} 
                
            \end{tabular}
            \caption{The performance boost (\%) of sequence matching performance (with sequence lengths of \textbf{K} = 2, 5, 10 and 15 images) in comparison to the single-frame-matching performance of all VPR techniques on the datasets mentioned in sub-section \ref{sequentialdatasets}}
            \label{performanceboostpercentage}
            \end{figure*}

        \subsection{Utilised VPR Techniques}
        In this work, sequence-based filtering is introduced into a number of state-of-the-art VPR techniques, namely HOG, CALC, AMOSNet, HybridNet and NetVLAD. Single-frame-based implementation of Zaffar \textit{et. al} \cite{zaffar2020vpr} is used for all 5 aforementioned VPR techniques. In Section \ref{results}, comparative results based on the above-mentioned performance metrics for these VPR techniques are presented along with discussion of the benefits and trade-offs of sequence-based filtering. 
        
        \subsection{Utilised Sequential Datasets} \label{sequentialdatasets}
        For this study, 4 sequential VPR datasets are used. The first dataset is Campus Loop dataset, which contains 100 query and 100 reference images. It poses challenges to any VPR system due to the high amount of viewpoint variation, seasonal variation and also the presence of statically-occluded frames. The second and third datasets are part of Gardens Point dataset \cite{sunderhauf2015performance} which contains both day and night images, that are divided as follows: 200 query images (day images) and 400 reference images (equally split into day images and night images). Nordland dataset \cite{nordland2013dataset} is the fourth dataset used which captures the drastic visual changes that seasonal variation can have on a place (spring, summer, autumn and winter). Since the most notable differences between seasons are seen during the summer and winter seasons, each VPR technique is tested here on the summer-to-winter traverses of the Nordland dataset. Fig. \ref{datasetimages} shows sample images taken from each dataset.

    \section{Results and Analysis}
    \label{results}

        This section presents the results for sequence-based filtering when used on top of the above-mentioned VPR techniques.
        
        \begin{figure*}
            \centering
            \begin{tabular}{ c c }

                \includegraphics[width=155pt]{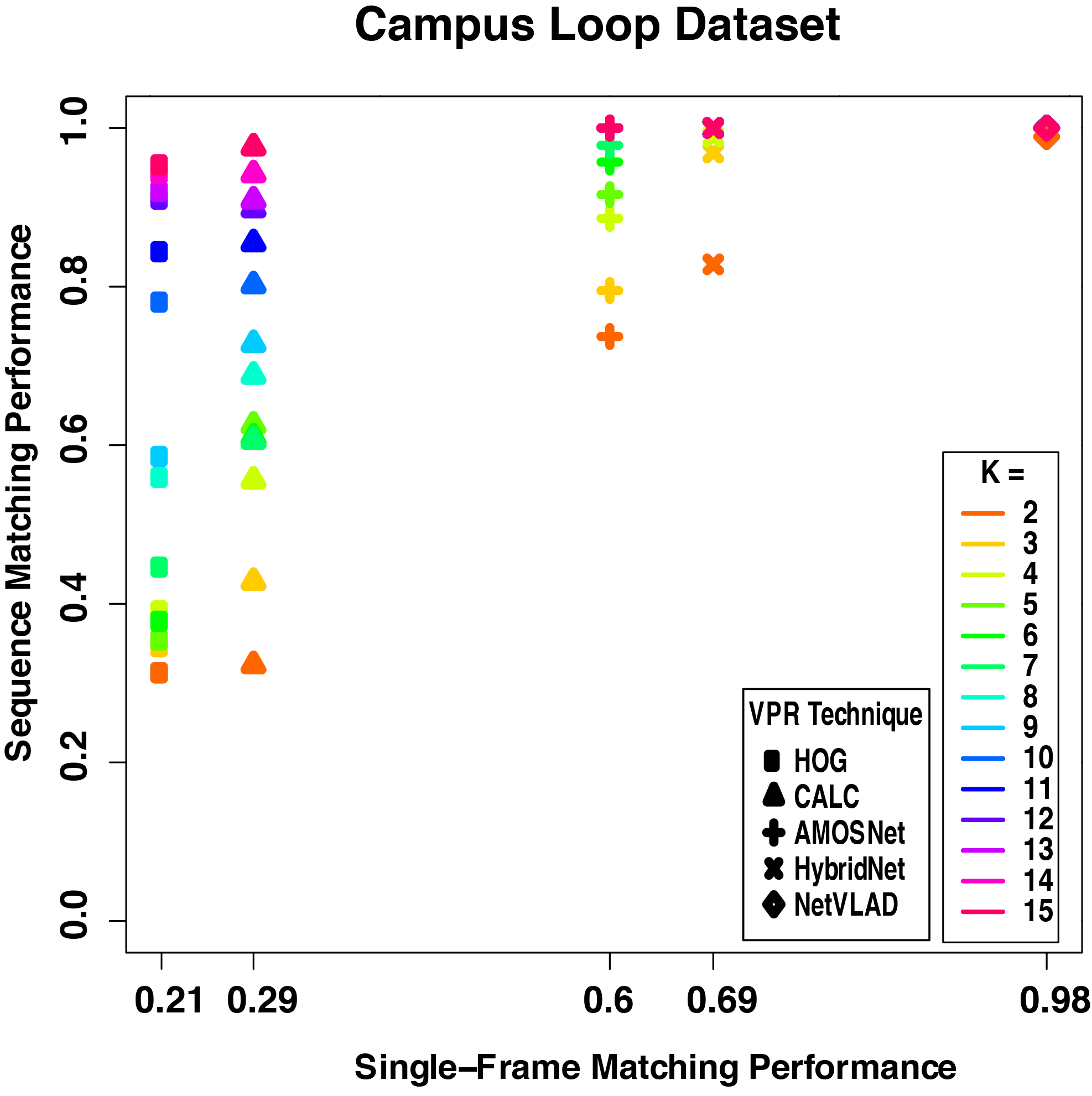} & 
                \includegraphics[width=155pt]{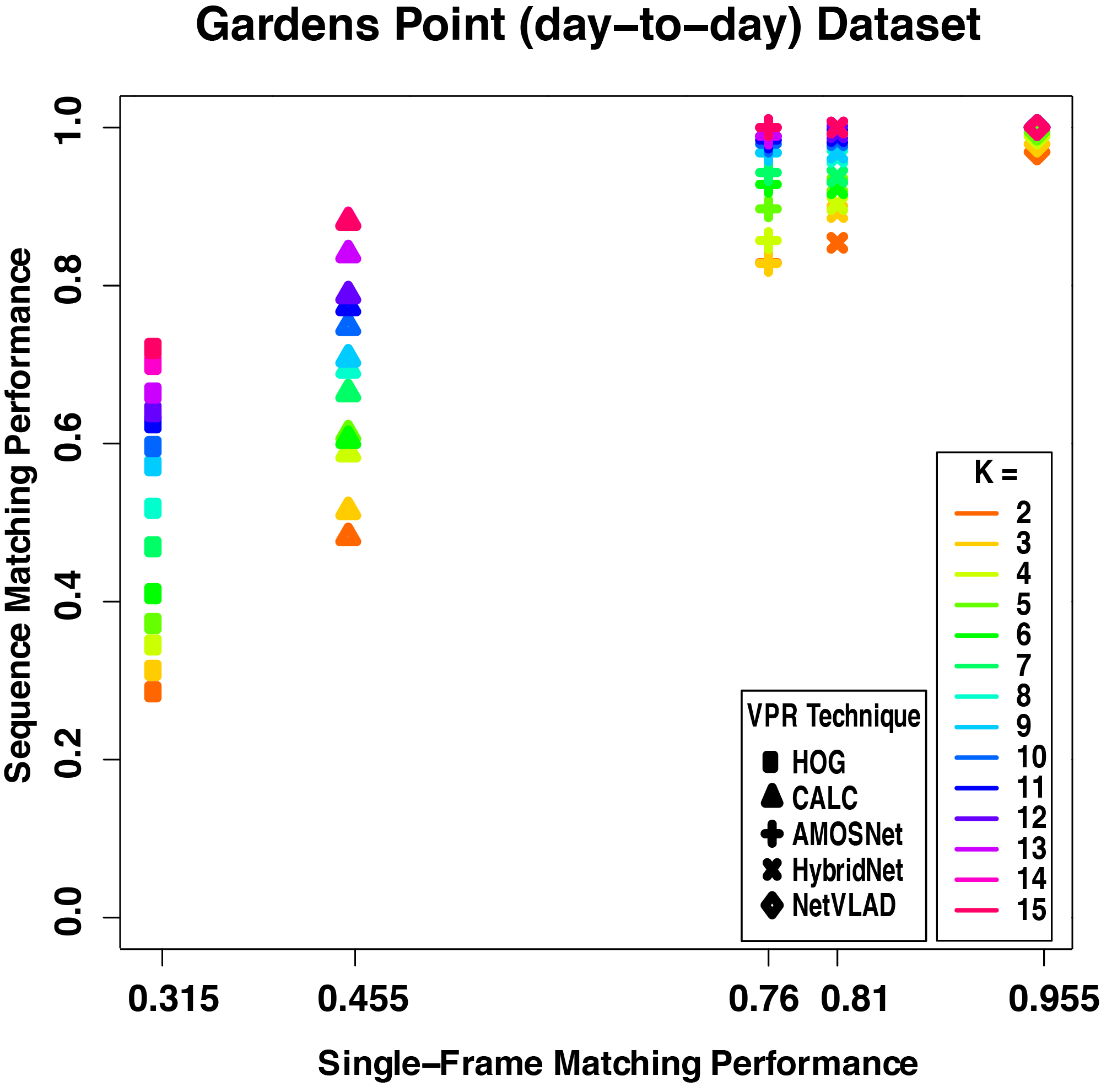} \\
                \includegraphics[width=155pt]{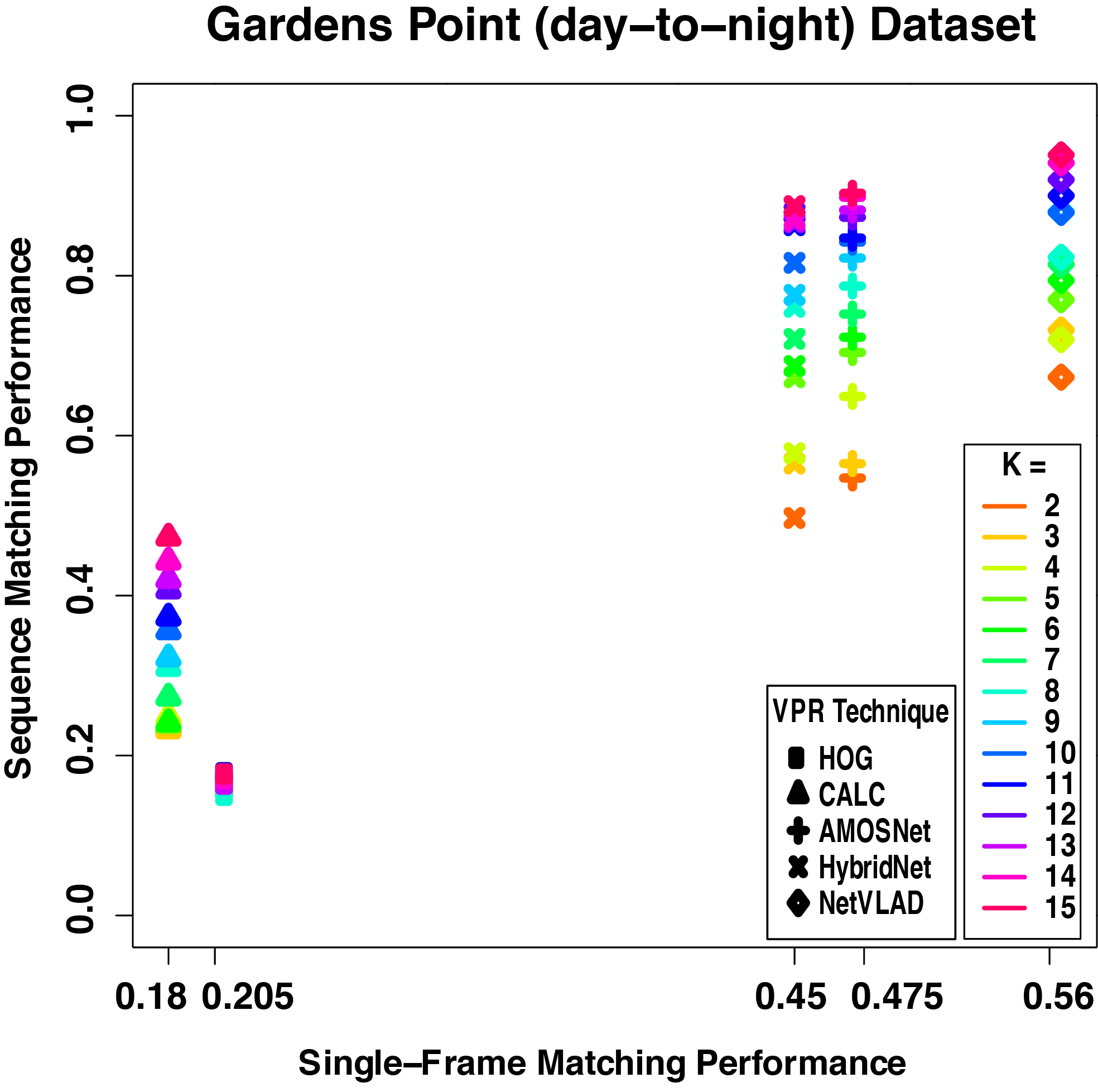}  &
                \includegraphics[width=155pt]{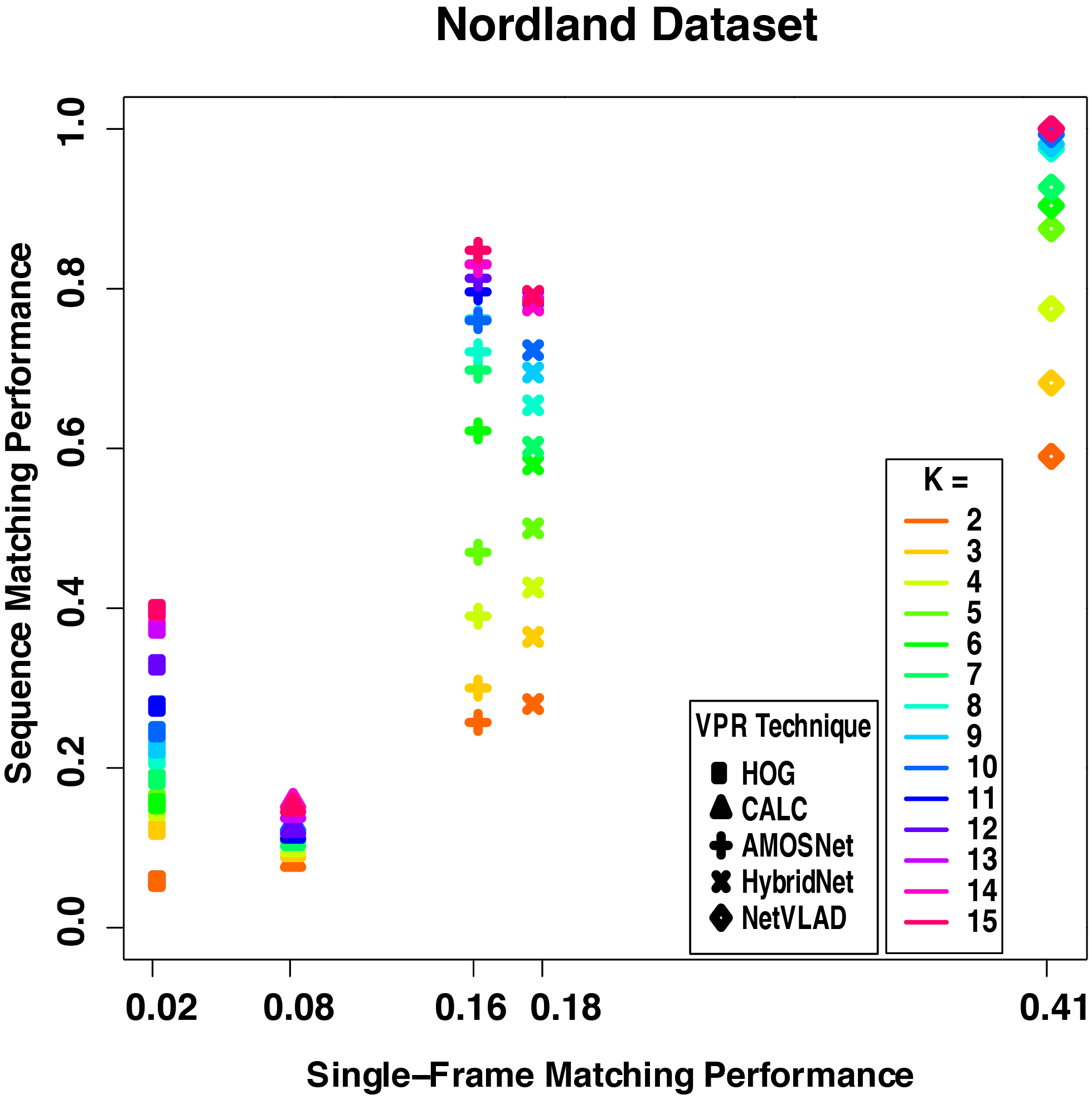}

            \end{tabular}
            \caption{The single-frame matching performance compared to the sequence matching performance for all 5 VPR techniques on all 4 datasets with sequence length up to 15 images.}
            \label{seqmatching}
            \end{figure*}

            \begin{figure*}
            \centering
            \begin{tabular}{ c c }

                \includegraphics[width=155pt]{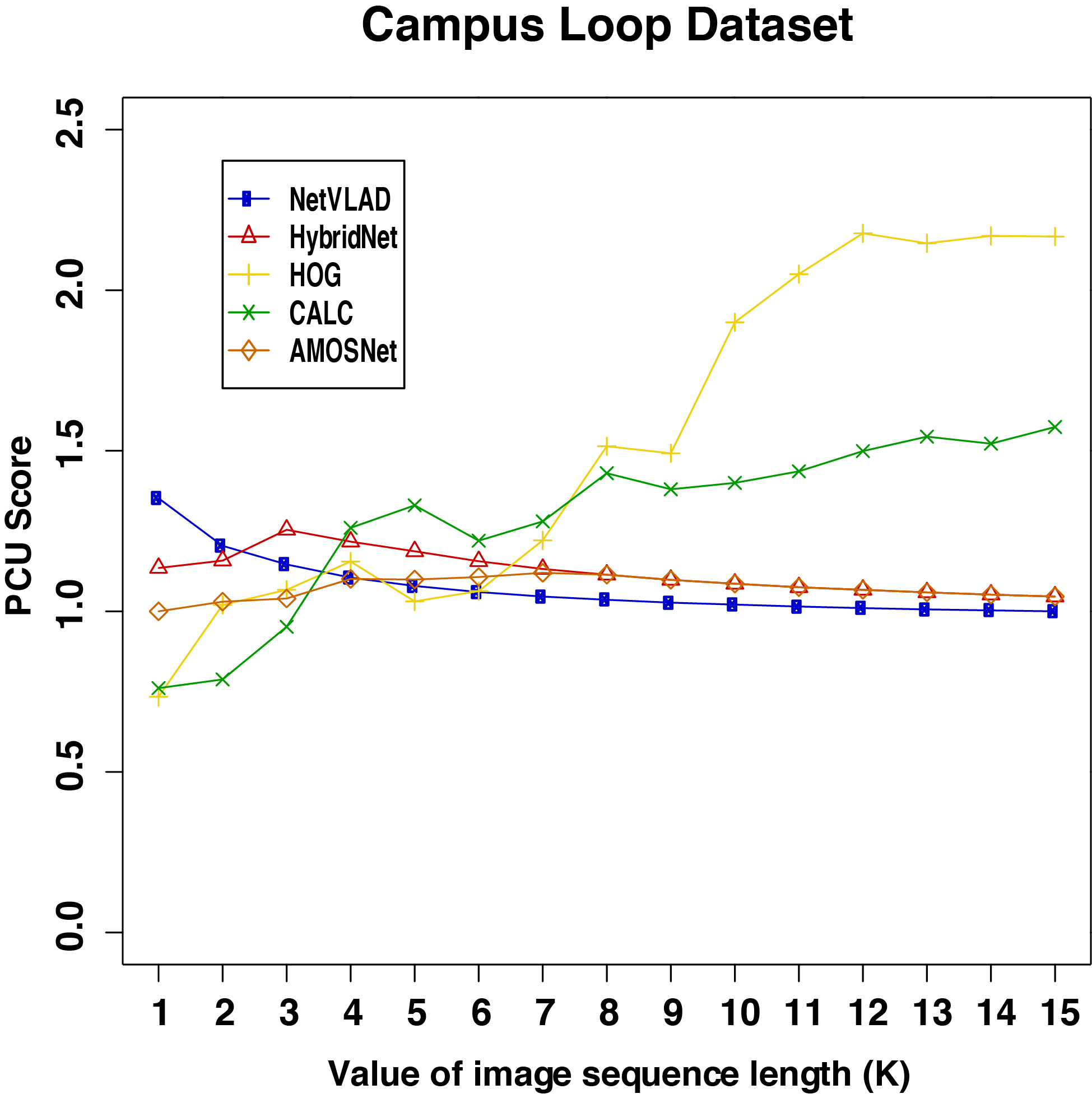} & 
                \includegraphics[width=155pt]{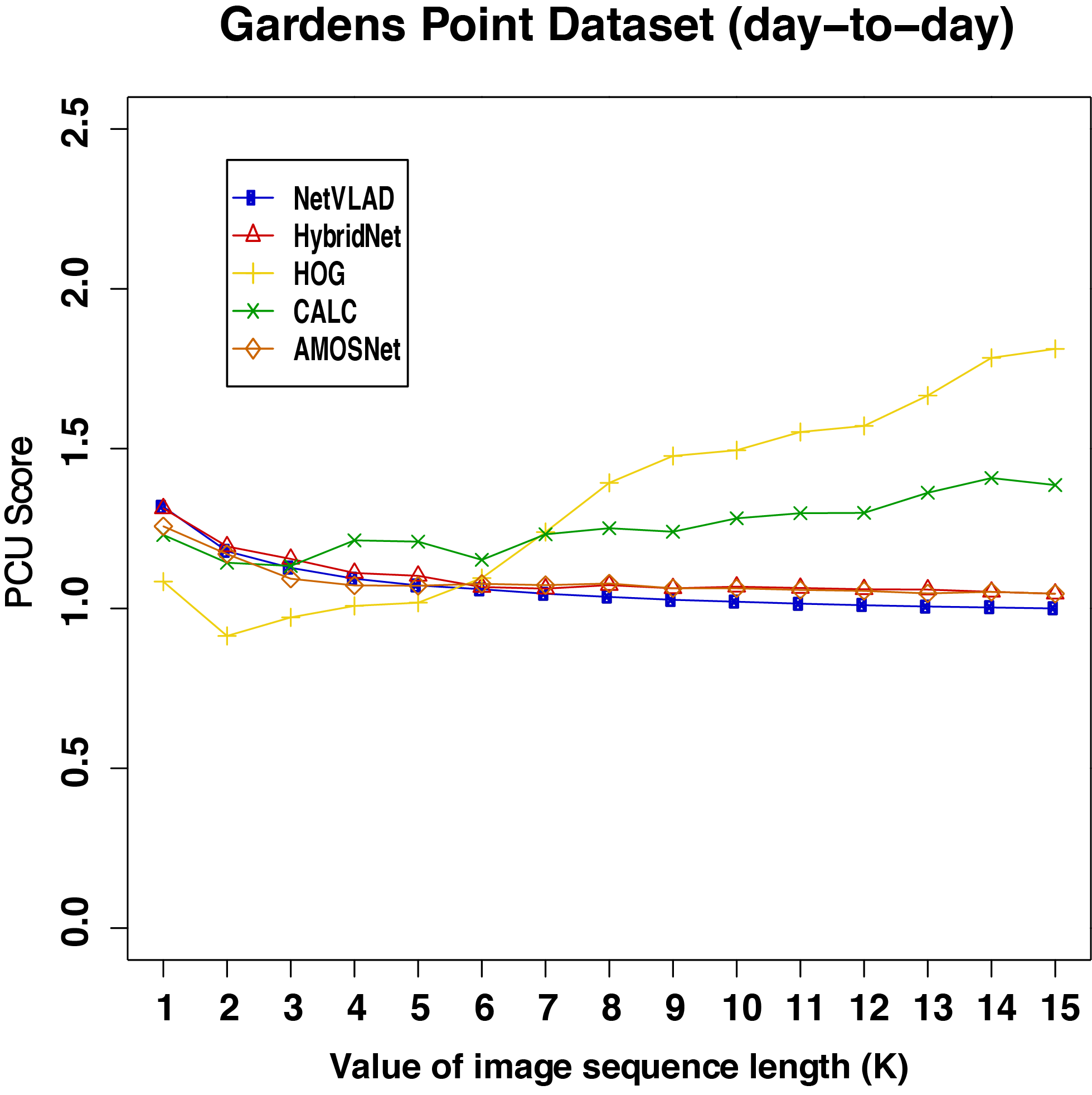} \\
                \includegraphics[width=155pt]{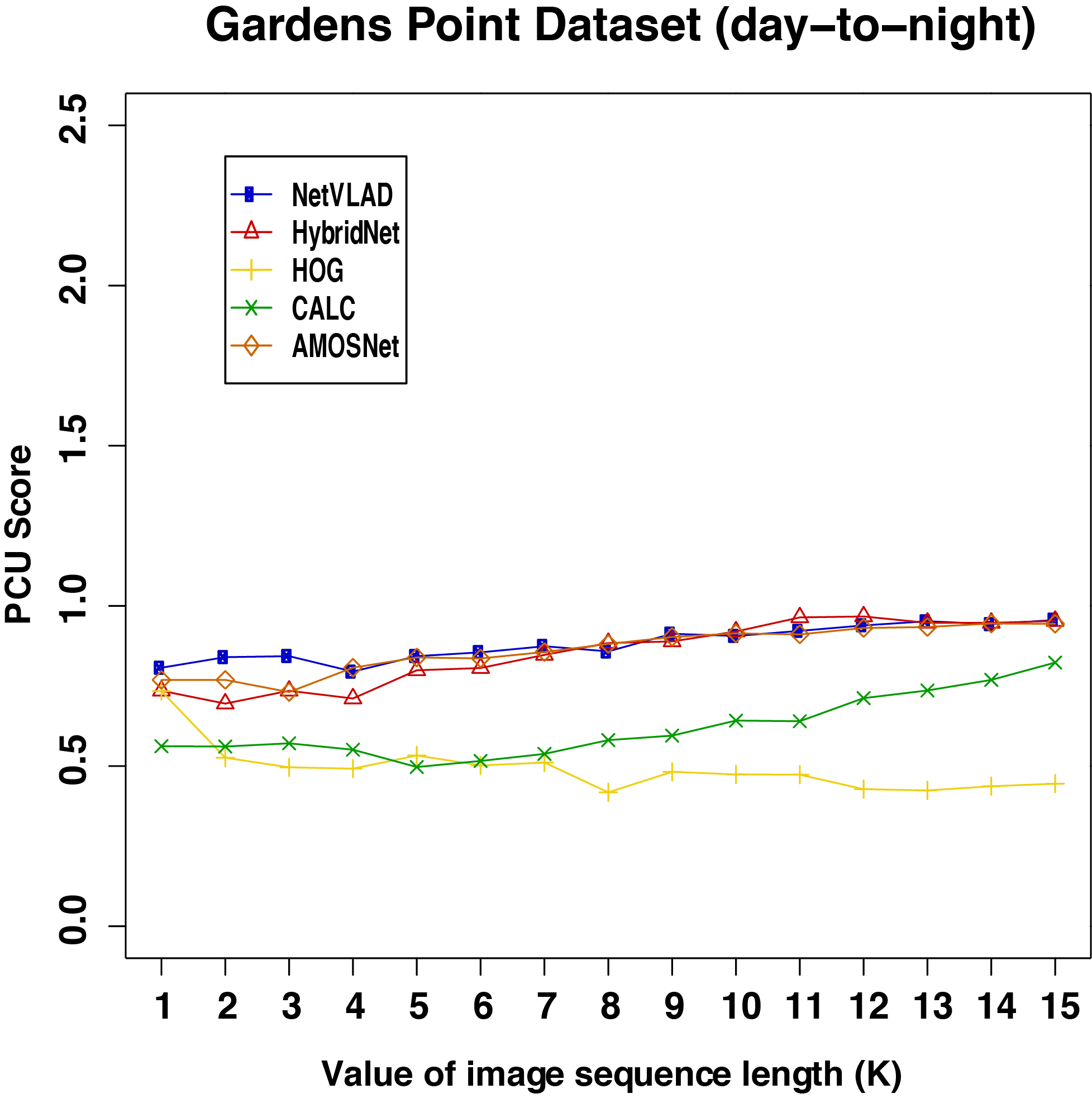}  &
                \includegraphics[width=155pt]{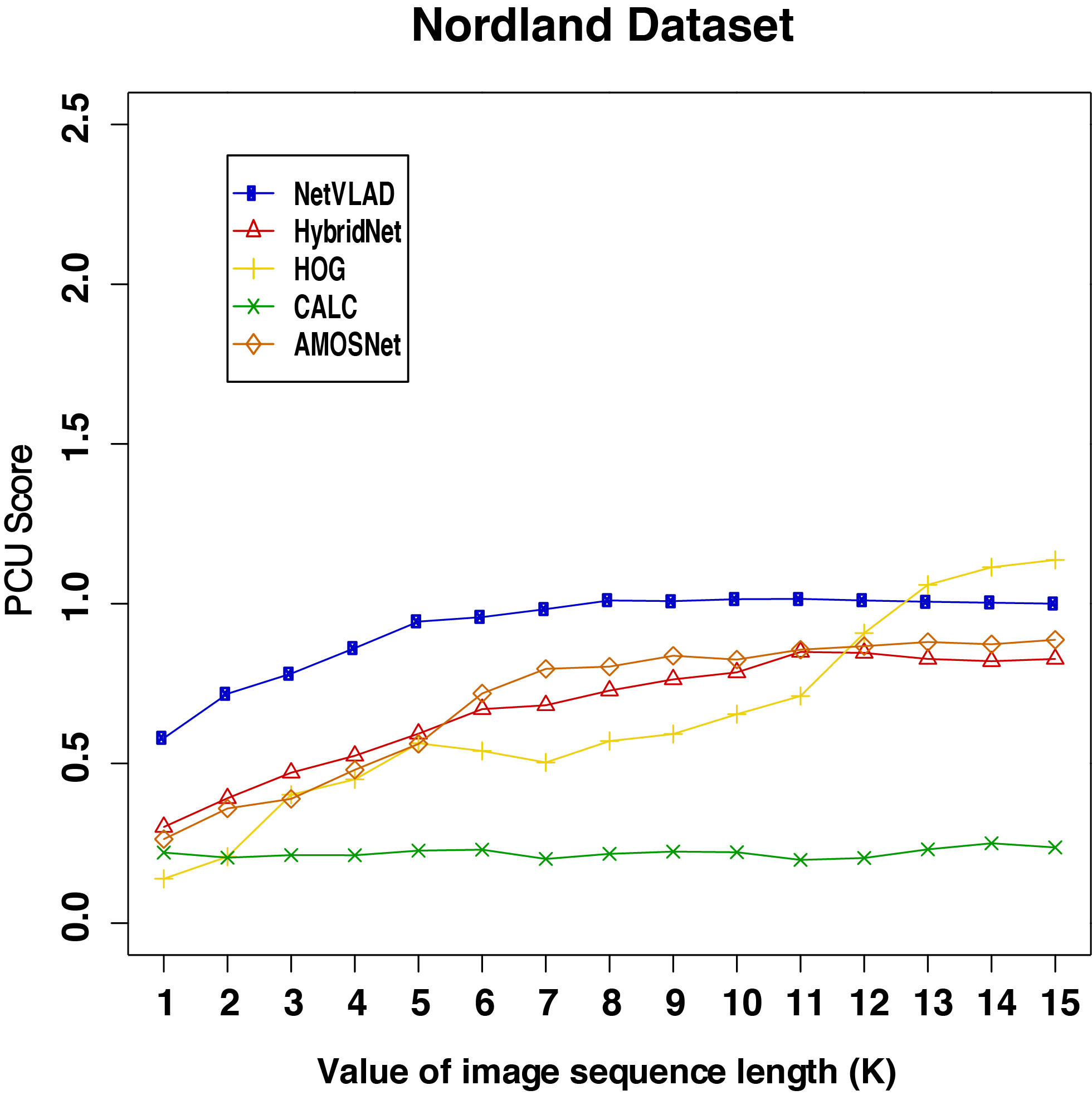} 
            \end{tabular}
            \caption{The PCU value of each VPR technique on each dataset with different sequence lengths. }
            \label{PCUseq}
            \end{figure*}

        \subsection{Place matching performance}    
            It is evident from Fig. \ref{performanceboostpercentage} that the addition of sequential filtering to a given single-frame-based VPR technique mostly improves the overall place matching performance of that technique. This suggests that by increasing the sequence length of a VPR technique, we will achieve better place matching performance. HOG achieves the highest performance boost (for sequence lengths of \textbf{K} = 10 and 15 images respectively) on all datasets except Gardens Point day-to-night dataset. VPR techniques such as AMOSNet and HybridNet have a substantial increase in performance on datasets such as Gardens Point day-to-night and Nordland while more simpler VPR techniques, such as CALC, achieve high performance boosts on Campus Loop and Gardens Point day-to-day datasets. However, VPR techniques which already achieve close-to-ideal matching performance, such as NetVLAD, do not benefit much from using an increased sequence length on certain datasets, such as on the Campus Loop and Gardens Point day-to-day dataset, where the performance boost of the system is negligible. This observation is important as using sequences instead of single images has computational drawbacks and should be avoided where unnecessary. We expand on this further in sub-section \ref{computationaleffects}.
            

             \subsection{Performance-boost Variations}
            \label{Performance_boost_Variations}
            In Fig. \ref{seqmatching}, we present the single-frame matching performance (x-axis) of each VPR technique in terms of accuracy and compare it with the sequence matching performance (y-axis) (2$\leq$\textbf{K}$\leq$15) of the same VPR techniques. Plotting the performance variations in this manner helps us to understand the amount of compression and expansion in performance boosts given sequence length variations for different VPR techniques, while also putting it on par with the single-image retrieval performance.\endgraf
            
            A common observation in existing literature has been that sequential-filtering mostly helps with introducing conditional-invariance \cite{milford2012seqslam}, however, we show (see results in Fig. \ref{seqmatching} on Gardens Point day-to-day dataset) that it also greatly helps in viewpoint-variant, conditionally-invariant scenarios. The performance boost of each VPR technique is directly linked to the intensity of conditional variations (and their effects on the scene appearance) in the dataset. The benefits of sequential-filtering are clearly enjoyed extensively by most techniques on datasets (Campus Loop and Gardens Point day-to-day) with less conditional changes than datasets (Nordland and Gardens Point day-to-night) with extreme conditional changes.\endgraf
            
            In contrast to the observations made above, the performance improvement of HOG is inconsistent on the Gardens Point day-to-night dataset, where the single-frame performance of this technique achieves similar or better place matching performance compared to that of the sequence matching performance. The presence of extreme viewpoint variation, illumination variation and also the presence of statically-occluded frames in the Gardens Point day-to-night dataset may affect the performance of this technique. Similarly, the improvement in the performance gained by using sequential-based matching for CALC is more limited when compared to other techniques on the Nordland dataset due to the presence of viewpoint and seasonal variation, as seen in Fig. \ref{seqmatching}. In such scenarios, evidently it is better for a system to switch to more sophisticated and invariant techniques, such as NetVLAD and HybridNet, even at the expense of higher computational needs. \endgraf
            
            In summary, some example cases where using a higher sequence length for a trivial VPR technique (such as HOG) is beneficial are laterally viewpoint variant and seasonally variant (but under similar illumination) scenes, e.g. driving a car in a different lane on a previously visited road in a different season. The increasing trend in performance of the HOG technique can be clearly seen in both Fig. \ref{performanceboostpercentage} and Fig. \ref{seqmatching}, for the Campus Loop and Gardens Point day-to-day datasets. However, for platforms that can have 3D or 6-DOF viewpoint changes, e.g. drones, UAVs etc, deep-learning-based techniques should be used instead of trivial techniques with high sequence length, which is also the case for highly illumination/conditionally variant scenes such as those found in the Gardens Point (day-to-night) and Nordland datasets. Our data supports the fact that deep-learning-based VPR techniques are better equipped to deal with these variations, and that they should be used in these scenarios instead of more simple VPR systems. Thus, we propose that having this prior knowledge can lead a system based on an ensemble of sequentially-filtered VPR techniques, which are switched accordingly dependent upon the environmental variation cues. This criteria will ensure that the most appropriate VPR technique is selected in each scenario, thus increasing the place matching performance, possibly at much lower computational costs as discussed in sub-section \ref{computationaleffects}.
        

        \subsection{Benefits and trade-offs of sequential filtering}
            This sub-section presents the benefits and trade-offs of sequential filtering while also answering key questions. 
            
        \subsubsection{Computational effects of sequential-filtering} 
        \label{computationaleffects}
            Due to the fact that we are matching sequences of images instead of the traditional single-frame approach, the feature encoding time for each VPR technique will be increased by \textbf{K} folds. Table \ref{table:encodingtimes} shows the feature encoding time of the 5 VPR techniques used in this work without sequential filtering. Because neural network-based VPR techniques, such as HybridNet, AMOSNet and NetVLAD already have increased feature encoding times, the addition of sequential filtering will lead to a drastic increase in processing time. Fig. \ref{PCUseq} shows the Performance-Per-Compute-Unit (PCU) of each VPR technique and the computational effects of using sequence lengths of up to 15 images. It is important to note that a significant increase in the PCU curves occurs when there is a notable increase in precision compared to the increase in encoding time. HOG achieves high PCU values due to both its low encoding times and high increase in precision when adding sequential filtering. \endgraf
            Apart from the computational downsides mentioned above, the latency in getting a match as it need to build up sequence has to be considered. Furthermore, shifting between two different routes that have not been traversed in that order in the map (switching latency) as well as the difficulties with variable velocities (solved partially with more sophisticated search or using odometry information) can lead to further computational constraints. This is especially important for resource constrained platforms as it may restrict its applicability in real world scenarios, due to the high amount of visual information that has to be processed.
        
            \subsubsection{Sequence-based filtering vs. single-image-based VPR}
             The data shows that for a VPR system that has poor performance on a dataset, the addition of sequence-based filtering may greatly improve its performance. Using a longer sequence length will have a higher impact in place matching performance. This is the case for HOG, which greatly benefits from the addition of sequence-based filtering. On the other hand, the single-image version of NetVLAD already achieves almost perfect results on both Campus Loop and Gardens Point (day-to-day) datasets and thus, the increased computational effects of sequential filtering for just a small gain in place matching performance may not evidently be desirable, as shown in Fig. \ref{PCUseq}. Empirically, increase of sequence length does not cause any reduction in the place matching performance but mostly yields better performance and therefore, given computational power, it may be desirable to use sequence-based techniques instead of single-image-based techniques.

            \subsubsection{Performance Benefits based on Sequential Filtering}
              As seen in both Fig. \ref{performanceboostpercentage} and Fig. \ref{seqmatching}, an increased sequence length for a given VPR technique will lead to higher performance on most datasets tested. However, different VPR techniques will require different sequence lengths depending on the performance of the system on a given dataset. When using sequence-based filtering, the boost in performance can be attributed to several reasons. Primarily, using an increased sequence length increases the chances of finding the best reference image for any given query image which also translates to reduced perceptual aliasing. The increased sequence length also improves the conditional-invariance of a VPR technique as shown by our results.
        
            \subsubsection{Light-weight vs. deep-learning-based VPR techniques blended with sequential-based filtering}
            It is evident that it is indeed possible to use a much simpler, light-weight VPR technique, paired with sequential filtering in order to match or even outperform the effectiveness of deep-learning-based VPR techniques on certain datasets. We have showed that the performance of a simpler VPR technique, such as HOG, can be drastically increased when using sequence-based filtering with a longer sequence length. The same can be said about CALC, which achieves good results when paired with sequential filtering. Moreover, both VPR techniques have a low feature encoding time, thus greatly benefiting from a PCU standpoint. Using the best VPR techniques (simpler systems with longer sequence lengths or deep-learning-based systems with smaller sequence length) for the right dataset will result in an overall better place matching performance, as discussed in sub-section \ref{Performance_boost_Variations}.

    \section{Conclusions}
     \label{conclusion}
    To bridge the gap of lack of a systematic study on sequence-based filtering for visual route-based navigation, this paper has conducted an in-depth investigation on the benefits and trade-offs of sequence-based filtering on top of single-frame-based VPR methods. This analysis is performed on 4 public sequential VPR datasets, that pose difficulties in place matching (appearance changes, viewpoint variations etc), using a variety of widely used performance metrics, such as Performance-per-Compute-Unit (PCU). Sequential filtering is introduced into a number of contemporary single-frame-based VPR methods in order to present the findings. The results show the effects of various sequence lengths on performance boost and suitable combinations of different VPR techniques and sequence lengths are determined, taking into consideration the computational effects of sequential-filtering, for the best place matching performance in different scenarios.

    {
    \small
    \bibliographystyle{ieeetr}
    \bibliography{root}
    }

\end{document}